\documentclass[preprint,12pt, a4paper]{elsarticle}

\usepackage{amssymb}
\usepackage{hyperref}
\usepackage{xcolor}
\usepackage{siunitx}
\usepackage{booktabs}

\journal{SoftwareX}

\begin{document}
\renewcommand{\labelenumii}{\arabic{enumi}.\arabic{enumii}}

\begin{frontmatter}


\title{WATCHED: A Web AI Agent Tool for Combating Hate Speech by Expanding Data}


\author[udc]{Paloma Piot}
\author[ind]{Diego Sánchez}
\author[udc]{Javier Parapar}
\address[udc]{IRLab, CITIC, Universidade da Coruña, Spain}
\address[ind]{Independent Researcher}
\address{paloma.piot@udc.es, diegosanchezlamas@gmail.com, javier.parapar@udc.es}

\begin{abstract}
Online harms are a growing problem in digital spaces, putting user safety at risk and reducing trust in social media platforms. One of the most persistent forms of harm is hate speech. To address this, we need tools that combine the speed and scale of automated systems with the judgement and insight of human moderators. These tools should not only find harmful content but also explain their decisions clearly, helping to build trust and understanding. In this paper, we present \textsc{WATCHED} a chatbot designed to support content moderators in tackling hate speech. The chatbot is built as an Artificial Intelligence Agent system that uses Large Language Models along with several specialised tools. It compares new posts with real examples of hate speech and neutral content, uses a BERT-based classifier to help flag harmful messages, looks up slang and informal language using sources like Urban Dictionary, generates chain-of-thought reasoning, and checks platform guidelines to explain and support its decisions. This combination allows the chatbot not only to detect hate speech but to explain why content is considered harmful, grounded in both precedent and policy. Experimental results show that our proposed method surpasses existing state-of-the-art methods, reaching a macro F1 score of 0.91. Designed for moderators, safety teams, and researchers, the tool helps reduce online harms by supporting collaboration between AI and human oversight.

\end{abstract}


\end{frontmatter}

\newpage

\section*{Metadata}
\label{}

\begin{table}[!ht]
\begin{tabular}{|l|p{6.5cm}|p{6.5cm}|}
\hline
\textbf{Nr.} & \textbf{Code metadata description} & \textbf{Metadata} \\
\hline
C1 & Current code version & v1.0 \\
\hline
C2 & Permanent link to code/repository used for this code version & \url{https://github.com/nulldiego/watched} \\
\hline
C3  & Permanent link to Reproducible Capsule & \url{https://github.com/nulldiego/watched}\\
\hline
C4 & Legal Code License   & MIT License \\
\hline
C5 & Code versioning system used & git \\
\hline
C6 & Software code languages, tools, and services used & python, pydantic, ollama, qdrant, transformers, torch \\
\hline
C7 & Compilation requirements, operating environments \& dependencies & Installation of dependencies: \texttt{uv} \\
\hline
C8 & If available Link to developer documentation/manual & NA \\
\hline
C9 & Support email for questions & paloma.piot@udc.es \\
\hline
\end{tabular}
\caption{Code metadata (mandatory)}
\label{codeMetadata} 
\end{table}

\section{Motivation and significance}

Hate Speech, defined as \textit{offensive, derogatory, humiliating, or insulting discourse \citep{fountahatespeech} promoting violence, discrimination, or hostility \citep{davidsonhatespeech} toward individuals or groups based on attributes such as race, religion, ethnicity, or gender \citep{hatelingo2018elsherief,ElSherief2018,hatemm2023}}, remains a serious social media issue, threatening individual safety and community cohesion \citep{GonzlezBailn2022}. Nearly one in three young people face cyberbullying \citep{KansokDusche2022}, and almost half of Black or African American adults report racial harassment online \citep{ADL2024}, underscoring the urgency of combating hate speech. Meanwhile, some platforms reduce investment in human moderation, prioritising engagement over consistent enforcement \citep{theguardianMetaFactcheckers}, leaving moderators under pressure and in need of automated tools for rapid, large-scale content analysis.

\begin{figure}[t]
  \centering
  \includegraphics[width=\linewidth]{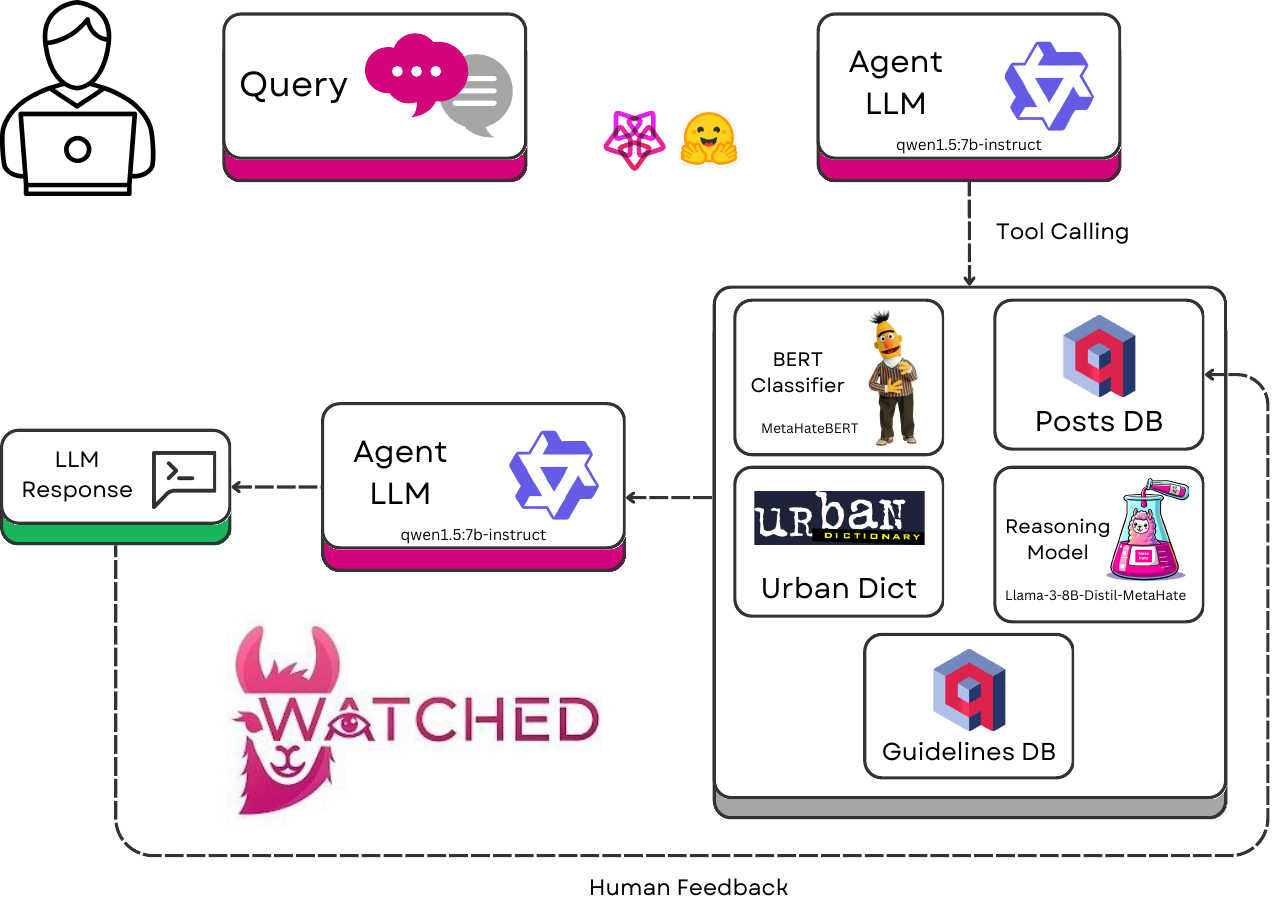}
  \caption{AI Agent pipeline including tools used during the classification task.}
  \label{fig:architecture}
\end{figure}

Hate speech detection has progressed from traditional machine learning with manual features \citep{waseem-hovy-2016-hateful, Davidson_Warmsley_Macy_Weber_2017, Chatzakou2017, Tahmasbi2018} to deep learning methods like CNNs and RNNs \citep{qian-etal-2019-benchmark}, and now to transformer models such as BERT and RoBERTa \citep{devlin-etal-2019-bert, grimminger-klinger-2021-hate, glavas-etal-2020-xhate}. Large Language Models (LLMs) are now state of the art, performing well in zero, few-shot and fine-tuning settings \citep{plaza-del-arco-etal-2023-respectful, roy-etal-2023-probing, wang2022toxicitydetectiongenerativepromptbased, piot-parapar-2025-decoding}. Tools exist for detecting explicit hate \cite{10.1145/3449280,ranasinghe-zampieri-2021-mudes}, analysing offensive content \cite{camacho-collados-etal-2022-tweetnlp,tillmann-etal-2023-muted,Le2023}, and debiasing datasets \cite{mosca-etal-2023-ifan}. One tool uses a BERT model for implicit hate classification, prompting LLMs for explanations \cite{damo:hal-04684950}.

Yet, these systems struggle with implicit language, slang, sarcasm, and context, and often lack interpretability.

To address these issues, we propose an AI agent to support human moderators in hate speech detection. This autonomous, task-driven software can perceive inputs, reason over context, and adaptively act for accurate responses.

Our contributions: (1) An AI Agent system for detecting hate speech on social networks; (2) integration of external knowledge, reasoning, and policies to assist moderators; (3) empirical evaluation against strong baselines, showing superior performance; (4) continuous hate speech data expansion and adaptation to new forms with human supervision.

\section{Software description}

We present \textsc{WATCHED} (\textbf{W}eb AI \textbf{A}gent \textbf{T}ool for combating \textbf{H}ate speech by \textbf{E}xpanding \textbf{D}ata), a demonstration tool for detecting, explaining, and augmenting hate speech data within an AI agent framework.

\subsection{Software architecture}

The system has a front-end and back-end (Figure~\ref{fig:architecture}). The front-end: (1) lets users input text for analysis, (2) shows the final classification with explanations and alignment to platform guidelines, (3) allows users to provide feedback on the correctness of the classification to expand data. The back-end orchestrates an AI agent that processes inputs, retrieves context, reasons, and generates final responses.

Unlike fixed pipelines, the agent reasons, adapts, and synthesises evidence by retrieving similar posts from a vector database, querying Urban Dictionary, invoking a BERT-based classifier, and calling a multi-step reasoning LLM. Tools are integrated via \texttt{pydantic\_ai}. The agent receives prompts, selects tools, gathers evidence, and synthesises a confident, explained decision.

\subsection{Software functionalities}

\subsubsection{Tools}

AI agents enhance LLMs by integrating external tools and APIs to overcome limitations like hallucinations and static knowledge \citep{rayhan2025llmenhancermergedapproach}. These agents invoke functions  when needed and incorporate the results into their reasoning, enabling real-time, dynamic interaction and decision-making. Moreover, they iteratively refine their prompts and adapt their behaviour based on the information they receive, allowing them to dynamically adjust their reasoning and improve task performance over time \citep{sapkota2025aiagentsvsagentic}. To enhance hate speech detection, our system includes:

\paragraph{Hate Speech Classifier}

\texttt{MetaHateBERT} \citep{Piot_Martin-Rodilla_Parapar_2024} is a transformer-based model fine-tuned for hate speech detection on various social platforms. It is based on BERT and trained on the MetaHate dataset \citep{Piot_Martin-Rodilla_Parapar_2024}. The model is designed for binary classification (hate vs. non-hate) and performs well on multiple benchmarks \citep{Piot2025}. We chose MetaHateBERT because it is accurate and fast, with a lightweight design that allows quick deployment and easy experimentation. MetaHateBERT can be accessed on Hugging Face at \url{irlab-udc/MetaHateBERT}.

\paragraph{Similar Posts}
\label{sec:similar-posts}

We included is Agentic Retrieval-Augmented Generation (RAG) system to fetch similar examples to the users' input query. At its core, RAG is about ``using an LLM to answer a user query, but basing the answer on information retrieved from a knowledge base'' \citep{NEURIPS2020_6b493230}. RAG offers grounded, domain-specific, and up-to-date responses with source transparency and control. However, it's limited by one-shot retrieval, query-document mismatch, shallow reasoning, and context size constraints. Agentic RAG overcomes these limitations by enabling iterative retrieval, optimised querying, deeper reasoning, and self-refinement for more accurate and robust responses \citep{huggingfaceAgentic}.

To create our RAG database, we selected the MetaHate dataset \citep{Piot_Martin-Rodilla_Parapar_2024} as the source of examples to store and retrieve. In total, we ingested \num{1164586} examples\footnote{We only ingested the train split of the MetaHate dataset, leaving the eval split unprocessed so it remains unseen by our AI agent and can be used for system evaluation.}. For storage, we used the Qdrant database \citep{qdrant} and computed embeddings with the \texttt{jinaai/jina-embeddings-v3} model \citep{jina-embeddings-v3-ecir25}. We chose Qdrant because, according to recent benchmarks, it offers high performance for multidimensional vector data \citep{Aumller2020}. The embedding model we selected ranks 17th on the MTEB English v2 leaderboard and 2nd among models with fewer than 1 billion parameters, with a memory footprint of 1092 MB. It supports over 89 languages and outperforms proprietary embeddings from OpenAI and Cohere on English tasks \citep{jinaJinaEmbeddings}. Its efficiency and multilingual support make it suitable for production environments. We computed the embeddings by encoding texts with the pretrained transformer model from Hugging Face, using ``text-matching'' task to produce 1024-dimensional vectors.

The tool encodes the user's input text into a semantic vector using our pretrained embedding model and then queries the Qdrant vector database containing the previously indexed posts with hate speech labels. It retrieves the top five most similar posts along with their similarity scores and hate speech flags. These retrieved examples serve as contextual evidence, enabling the AI agent to enhance its reasoning and generate more informed responses.

\paragraph{Urban Dictionary}

To address limitations in LLMs' understanding of informal or evolving language, we include a tool that queries Urban Dictionary \citep{urbandictionary} for user-specified terms. This is motivated by two main factors. First, language on social media evolves rapidly, and LLMs, or any other machine learning model, may lack up-to-date knowledge of newly coined slang or culturally-specific expressions \citep{Panjaitan2024, Bushra2025}. Urban Dictionary, as a crowd-sourced and constantly updated resource, can provide informal definitions that supplement the model's static knowledge. Second, prior research shows that models often over-predict hate speech when exposed to non-standard dialects, slang, or African American Vernacular English (AAVE), conflating linguistic style with toxicity \citep{sap-etal-2019-risk, davidson-etal-2019-racial}. By explicitly retrieving and injecting definitions of unfamiliar or stigmatised terms into the reasoning process, the tool helps mitigate false positives driven by unfamiliarity with informal language.

\paragraph{Reasoning} 

To support contextual reasoning during hate speech classification, we integrate a tool that invokes a distilled large language model fine-tuned for this task. Specifically, the tool takes a natural language prompt, refined by the agent, and a set of retrieved similar posts (including the texts, their similarity score, and the hate speech flags), and uses them as inputs to generate a reasoned judgement. Moreover, the agent might decide to include in the prompt information from other tools such as MetaHateBERT classification and confidence score and Urban Dictionary's definitions. The underlying selected model, \texttt{Llama-3-8B-Distil-MetaHate} (from now onwards, \texttt{Distil MetaHate}), was trained on the MetaHate dataset\footnote{Note that it was not trained on our test split.} and distilled for efficiency and interpretability \citep{Piot2025}. It enables the system to integrate retrieved evidence with reasoning to generate justified decisions, overcoming the limitations of purely similarity-based, rule-based, or binary classification approaches. This tool helps the system explain its decisions more clearly by combining context with relevant external knowledge.

\paragraph{Social network guidelines}

To enhance the interpretability and traceability of hate speech classification, we integrate a tool that incorporates content moderation guidelines from major social networks. This component supplements the reasoning model's output by grounding its explanations in established policy definitions.

Social media platforms typically enforce strict rules against harmful content, including hate speech, harassment, and incitement to violence. These rules are formalised in their content policies, which provide detailed definitions and examples of prohibited language. By referencing these official guidelines, our system adds a layer of normative context to its classifications, enabling users to understand why a particular post may be deemed harmful according to platform-specific standards.

We compiled hate speech guidelines from major platforms including \href{https://support.reddithelp.com/hc/en-us/articles/360045715951-Promoting-Hate-Based-on-Identity-or-Vulnerability}{Reddit}, \href{https://help.x.com/en/rules-and-policies/hateful-conduct-policy}{X} (formerly Twitter), and \href{https://transparency.meta.com/en-gb/policies/community-standards/hateful-conduct/}{Meta} (covering Facebook and Instagram), as well as institutional definitions from \href{https://www.un.org/en/hate-speech/}{UNESCO} and the \href{https://www.un.org/en/hate-speech/}{United Nations}. The guidelines are stored in a vector database, but unlike the setup described in Section~\ref{sec:similar-posts}, we used \texttt{crawl4ai} to scrape the guidelines in markdown format. We then applied a chunking algorithm and prompted \texttt{qwen2.5:7b-instruct- q4\_K\_M} to generate a title and summary for each chunk. The resulting titles were used to compute the embeddings stored in the database. When the AI agent produces a decision and rationale, we retrieve the most relevant policy snippets to support or contextualise the model's explanation, linking detected hate constructs with established platform or institutional definitions.

\begin{figure}[t]
  \centering
  \includegraphics[width=0.6\linewidth]{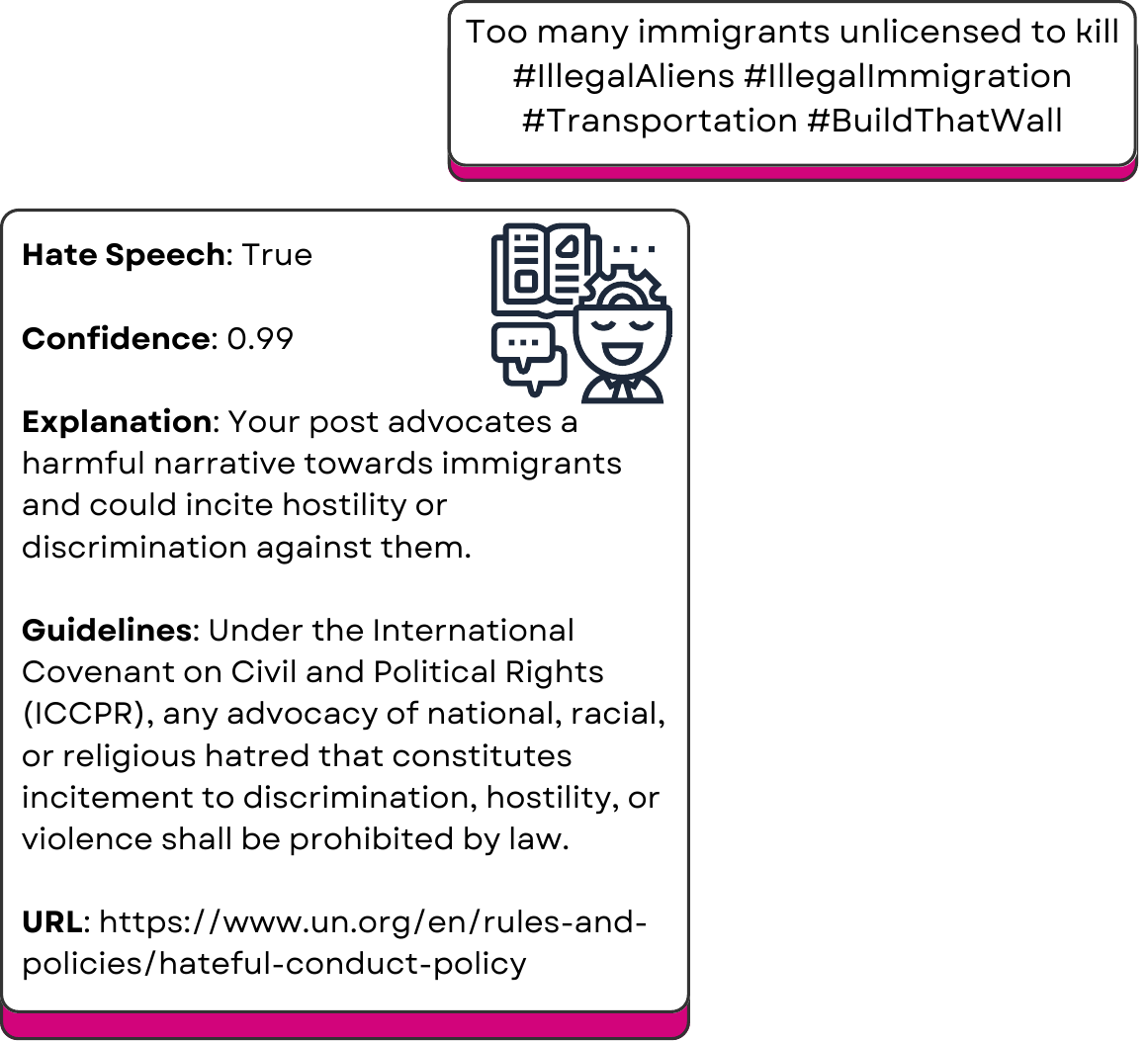}
  \caption{Social network guidelines example.}
  \label{fig:guidelines}
\end{figure}

This alignment not only improves the transparency of the agent's decisions but also supports real-world applicability, especially in moderation pipelines where compliance with platform policies is essential.

\subsubsection{Other features}

\paragraph{Data augmentation}

At the final stage of the pipeline, the user interface presents the system's output, consisting of a classification label, an explanation justifying the decision, and relevant social network guidelines supporting the outcome. After reviewing this information, users are prompted to provide feedback by either confirming the classification as correct or flagging it as incorrect. This feedback mechanism directly contributes to a dynamic refinement process. When feedback is submitted, the RAG database is updated with the labelled example: if the user confirms the prediction, the model's label is stored; if rejected, the opposite label is recorded. This updated data point is added to the system's evolving knowledge base.

This functionality promotes continuous learning by incorporating human-in-the-loop validation into the model's reasoning process. It helps prevent the system from remaining static, enabling it to adapt over time to evolving language patterns, domain shifts, and edge cases that may not have been present in the original training data.

\subsection{System Evaluation}

\subsubsection{Dataset}

We selected MetaHate dataset \citep{Piot_Martin-Rodilla_Parapar_2024}, a dataset comprising more than 1.2 million English labelled hate speech posts collected from 36 different datasets. For evaluating the system, we used the same subsample of MetaHate  that was employed in the \texttt{Distil MetaHate} benchmarks \citep{Piot2025}. We reannotated this sample to improve the reliability of our results. Two domain experts independently annotated the data. In cases of disagreement, they discussed the message and reached a consensus on the final label.\footnote{The new labels are released in our repository, together with the original labels.} This ensures comparability with prior work and maintains consistency across evaluation settings. Importantly, none of the examples used in our evaluation are part of the RAG database, guaranteeing a clean separation between training and external knowledge access. The evaluation sample contains \num{2001} instances.

\subsubsection{Models}

\paragraph{Agent LLMs} We selected \texttt{qwen2.5:7b-instruct-q4\_K\_M} as the agent model because of its small size and efficiency, which make it suitable for interactive chatbot settings. In this work, we focus on a single agent LLM to power our demonstration system. A comparative evaluation of alternative agent LLMs is considered out of scope, as our aim is to showcase the integration and functionality of a full hate speech analysis agent rather than benchmark agent models.

\paragraph{Baselines} We compare our AI agent against a set of baselines, selected to reflect different modelling strategies.

\begin{itemize}
    \item \texttt{MetaHateBERT} \citep{Piot_Martin-Rodilla_Parapar_2024}: A BERT-based model fine-tuned for hate speech detection. We include it to assess the benefit of replacing a standalone classifier with our multi-tool agent setup.
    \item \texttt{Distil MetaHate} \citep{Piot2025}: A distilled, specialised LLM used within our reasoning tool. Comparing the full agent to this single model isolates the impact of combining multiple reasoning and classification components.
    \item \texttt{Llama 3 8B}: A general-purpose LLM that served as the student model for the distilled variant above. This baseline shows the difference between fine-tuning for hate speech and using a general LLM.
    \item \texttt{Llama 3 70B}: The original teacher model used to distill \texttt{Distil MetaHate}. It serves to evaluate the trade-offs between model size and performance in our agent compared to large-scale LLMs.
    \item Perspective API \citep{Lees2022}: A widely adopted production tool that outputs a toxicity score. We use it as a benchmark for real-world applicability, binarising its output with a 0.5 threshold.
\end{itemize}

\subsubsection{Experimental settings}

We adopt the baselines and experimental settings from \cite{Piot2025}, and we report the metrics on the reannotated test split. For our agent, we use the \texttt{pydantic\_ai} framework, defining an \texttt{Agent}, configured with a custom system prompt\footnote{The full prompt is available on our GitHub.}, our set of tools, and an expected output format that includes a binary label, confidence score, and explanation. The agent is allowed up to five retries per query, with instrumentation enabled for traceability.

\subsubsection{Results}

Table \ref{tab:experiments} presents a comparison between our proposed system and the baselines. \textsc{WATCHED} achieves the highest performance across all F1 metrics, with an F1 scores surpassing 0.91. These results substantially outperform both the Llama 3 models (8B and 70B) in Few-Shot CoT settings and specialised models like \texttt{MetaHateBERT} and \texttt{Distil MetaHate}, the latter of which previously led the benchmarks with F1\textsubscript{MACRO} scores of 0.8801 and 0.8807 respectively. Additionally, the Perspective API lags behind with a F1\textsubscript{MACRO} of 0.7985, highlighting its limitations in this evaluation setting. These findings underscore the effectiveness of our agent-based approach in achieving high precision in hate speech detection.

\begin{table}
  \centering
  \label{tab:experiments}
  \begin{tabular}{l rrr}
    \toprule
    & F1 & F1\textsubscript{MICRO} & F1\textsubscript{MACRO} \\
    \midrule
    \texttt{Llama 3 8b} \textit{(Few-Shot CoT)} & 0.8219 & 0.8221 & 0.8152 \\
    \texttt{Llama 3 70b} \textit{(Few-Shot CoT)} & 0.8542 & 0.8536 & 0.8496 \\
    \texttt{Distil MetaHate} \citep{Piot2025} & 0.8826 & 0.8816 & 0.8807 \\
    \texttt{MetaHateBERT} \citep{Piot_Martin-Rodilla_Parapar_2024} & 0.8836 & 0.8831 & 0.8801 \\
    \texttt{Perspective API} \citep{Lees2022} & 0.8061 & 0.8066 & 0.7985 \\
    \midrule
    \textsc{WATCHED} & \textbf{0.9168} & \textbf{0.9165} & \textbf{0.9139} \\
    \bottomrule   
    \end{tabular}
  \caption{Results on the reannotated eval subset.}
\end{table}

\subsubsection{Ablation study}

To understand the individual contribution to the overall performance of the AI agent tools, we decided to perform an ablation study.  The goal is to assess the impact and importance of each tool integrated into the agent's architecture. To do so, we systematically remove individual tools from the full setup and measure the resulting performance. The following configurations were evaluated:

\begin{itemize}
    \item Agent LLM with all tools (\textsc{WATCHED}).
    \item Agent LLM without any tools
    \item Agent LLM without \textit{Hate Speech Classifier}
    \item Agent LLM without \textit{Similar Posts}
    \item Agent LLM without \textit{Urban Dictionary}
    \item Agent LLM without \textit{Reasoning}
\end{itemize}

Note that we do not ablate the social network guidelines component, as it does not directly affect classification decisions but rather provides explanatory context and traceability.

\begin{table}
  \centering
  \label{tab:ablation}
  \begin{tabular}{l rrr}
    \toprule
    & F1 & F1\textsubscript{MICRO} & F1\textsubscript{MACRO} \\
    \midrule
    no tools & 0.8148 & 0.8184 & 0.8053 \\
    w/o \textit{Hate Speech Classifier} & 0.8608 & 0.8598 & 0.8571 \\
    w/o \textit{Similar Posts} & 0.8786 & 0.8789 & 0.8733 \\
    w/o \textit{Urban Dictionary} & 0.8784 & 0.8780 & 0.8753 \\
    w/o \textit{Reasoning} & 0.8887 & 0.8882 & 0.8853 \\
    \midrule
    \textsc{WATCHED} & \textbf{0.9168} & \textbf{0.9165} & \textbf{0.9139} \\
    \bottomrule   
    \end{tabular}
  \caption{Ablation study results.}
\end{table}

Table \ref{tab:ablation} shows the results of the ablation study. The full model achieves the highest performance across all metrics, confirming the benefit of incorporating multiple external tools. Removing all tools causes a substantial drop in performance (F1\textsubscript{MACRO}: 0.8053), showing that external knowledge sources are critical to the agent's effectiveness.

Among individual tools, \texttt{MetaHateBERT}, the \textit{Hate Speech Classifier} tool, contributes the most; removing it results in the largest drop (F1\textsubscript{MACRO}: 0.8571) among single-tool ablations. This suggests that domain-specific language models provide strong semantic cues for hate speech classification.

The removal of \textit{Similar Posts} and \textit{Urban Dictionary} also leads to noticeable performance drops (F1\textsubscript{MACRO}: 0.8733 and 0.8753, respectively), highlighting the importance of contextual grounding and slang understanding. 

Notably, removing the \textit{Reasoning} tool, which only disables the external call to \texttt{Distil MetaHate} while preserving the agent's own internal reasoning, results in a moderate performance decrease (F1\textsubscript{MACRO}: 0.8853). This suggests that while the LLM can still perform reasoning independently, the distilled rationale model provides additional structured insight that improves classification consistency and quality.

\begin{figure}[t]
  \centering
  \includegraphics[width=\linewidth]{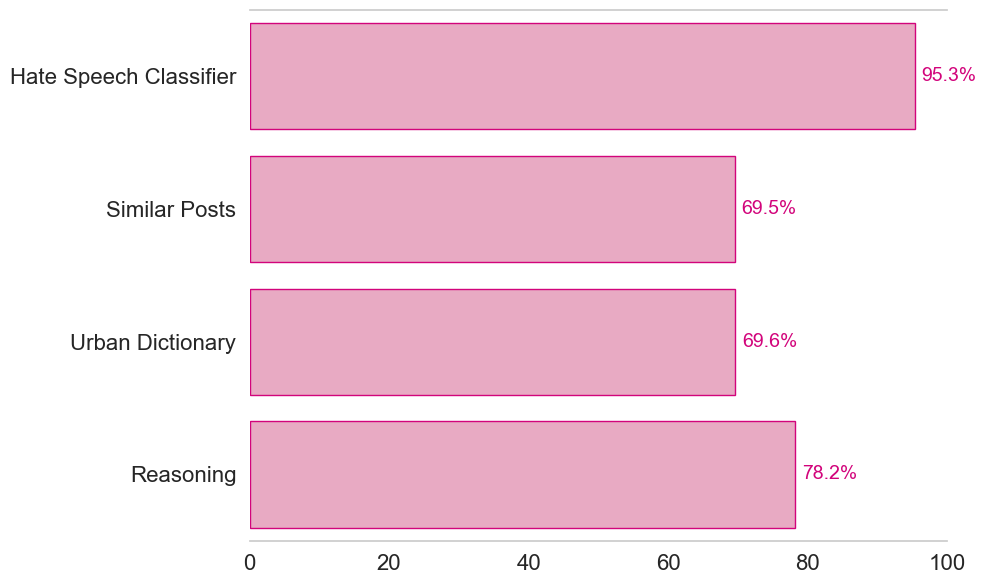}
  \caption{Percentage of inputs invoking each tool.}
  \label{fig:tool_percentage}
\end{figure}

Figure \ref{fig:tool_percentage} shows the percentage of inputs that called each tool at least once. The \textit{Hate Speech Classifier} was used in \num{95.3}\% of cases, making it the most commonly invoked tool. The \textit{Similar Posts} and \textit{Urban Dictionary} tools were each used in about \num{69.5}\% and \num{69.6}\% of queries, respectively. The \textit{Reasoning} tool appeared in \num{78.2}\% of inputs, indicating its frequent role in deeper analysis. These results highlight the widespread use of the classifier, with other tools supporting the process in many, but not all, cases.

\begin{figure}[t]
  \centering
  \includegraphics[width=\linewidth]{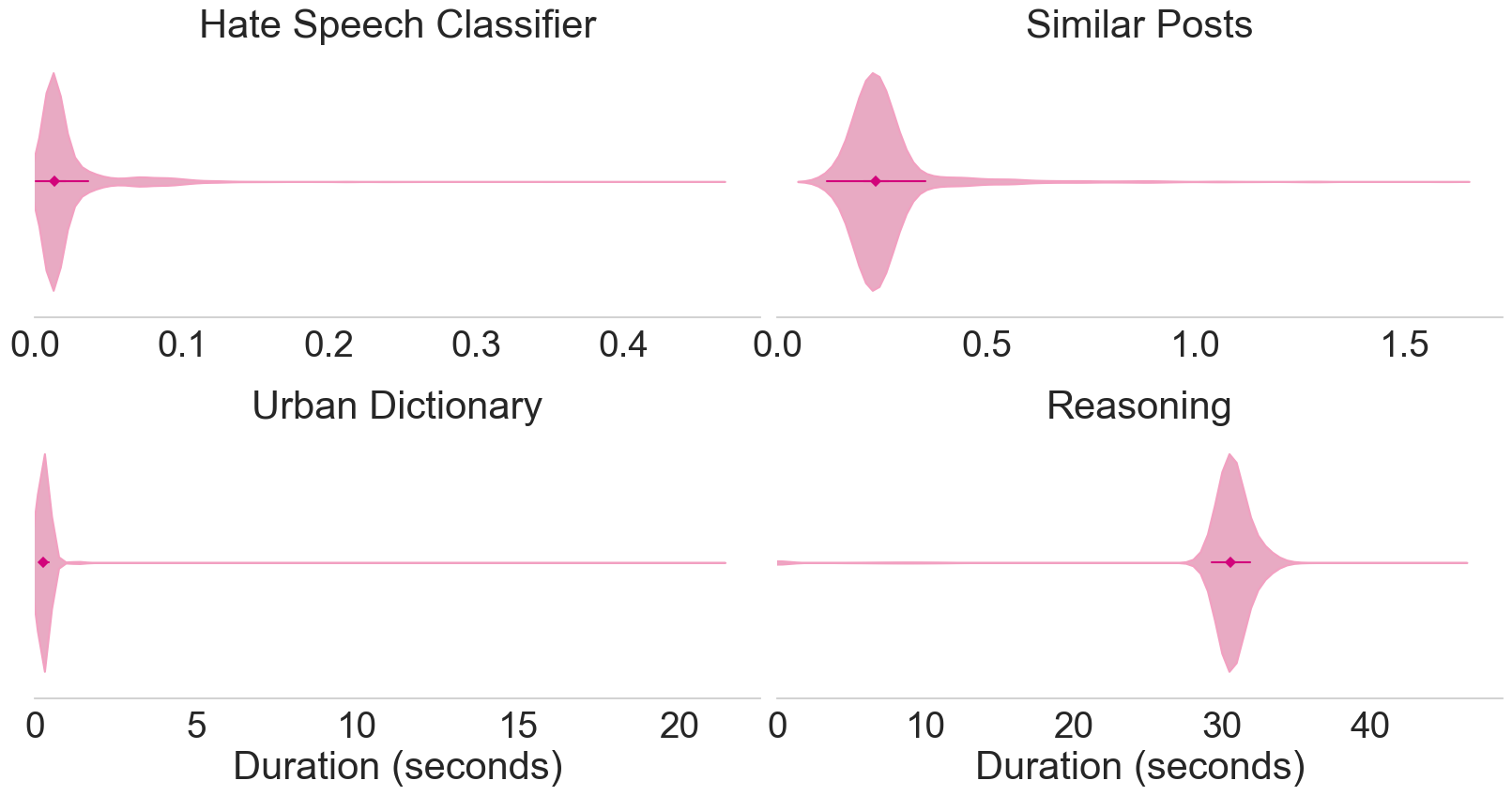}
  \caption{Distribution of tool execution duration.}
  \label{fig:tool_duration}
\end{figure}

Figure \ref{fig:tool_duration} presents the distribution of tool execution times. The \textit{Hate Speech Classifier}, \textit{Similar Posts}, and \textit{Urban Dictionary} tools exhibit fast response times, each completing in under one second, facilitating prompt initial evaluations. The \textit{Hate Speech Classifier} is the fastest, with mean responses below \num{0.1} seconds, followed by the \textit{Similar Posts} tool, where mean values are around \num{0.25} seconds. The call to \textit{Urban Dictionary} has mean values around \num{1} second. By contrast, the \textit{Reasoning} tool (which called \texttt{Distil MetaHate} model) requires an average of \num{30} seconds per query, reflecting the greater computational complexity involved in deeper analytical processing. Together, these findings highlight the system's strategic balance between efficient detection and comprehensive reasoning.

\section{Illustrative examples}

\begin{figure}[t]
  \centering
  \includegraphics[width=0.8\linewidth]{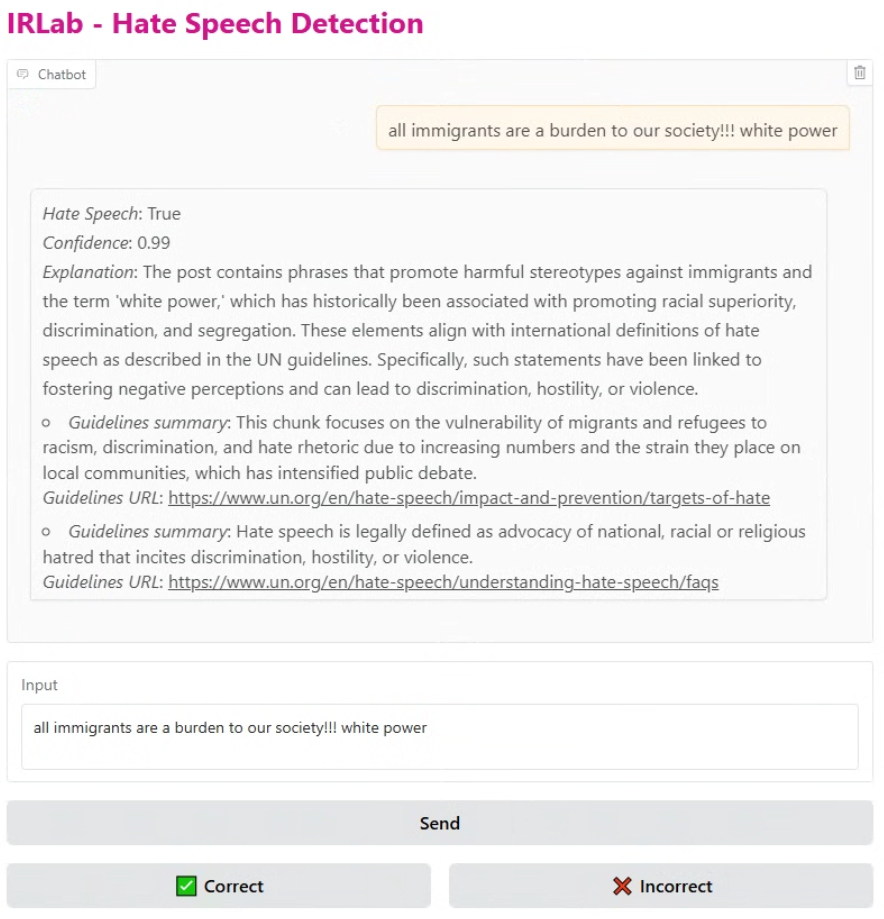}
  \caption{Illustrative example of the chatbot UI. The bottoms below allow the user to provide feedback on the accuracy of the response, and store the new message with the correct label from the human feedback.}
  \label{fig:example}
\end{figure}

Figure \ref{fig:example} shows an example with human-in-the-loop feedback for data expansion.

\section{Impact}

\textsc{WATCHED} opens promising research avenues in computational social science, NLP, and human–AI collaboration. By combining RAG, hate speech detection, and policy-grounded explanations, it enables investigation into how AI can better address implicit, slang-heavy, and context-dependent hate speech: areas where prior systems often fail. Human-in-the-loop feedback supports studies on the evolution of hateful discourse, the effectiveness of interventions, and AI adaptability to new linguistic and cultural trends. Integrating platform guidelines in outputs also allows comparisons between institutional definitions of harmful content and real-world moderation practices.

The system can be integrated into moderation workflows, helping moderators detect, explain, and document harmful content with greater transparency. Its open-source, modular design supports academic research, demonstrations, and adaptation to other online harms. These characteristics create opportunities for adoption by NGOs, safety labs, and commercial platforms seeking interpretable, adaptable moderation tools.

\section{Conclusions}

In this work we present \textsc{WATCHED}, an AI Agent system conceived to support social media networks moderators on detecting and understanding hate speech. The tool allows the moderators input a social network message and get a detailed response on if the input constitutes hate speech or note, and the explanation, together with the selected target social network guidelines adherence. To reach that conclusion, the agent has several tools under its control and decides what to execute to achieves the most optimal result. Our system outperforms current state-of-the-art methods and provides clear, understandable decisions to help human moderators.

\section*{Computational resources}

Experiments were conducted using a private infrastructure, which has a carbon efficiency of 0.432 kgCO$_2$eq/kWh. A cumulative of 480 hours of computation was performed on hardware of type GTX 1080 (TDP of 180W). Total emissions are estimated to be 37.32 kgCO$_2$eq of which 0 percents were directly offset. Estimations were conducted using the \href{https://mlco2.github.io/impact#compute}{MachineLearning Impact calculator} presented in \cite{lacoste2019quantifying}.

\section*{Acknowledgements}
\label{}
The authors thank the funding from the Horizon Europe research and innovation programme under the Marie Skłodowska-Curie Grant Agreement No. 101073351. Views and opinions expressed are however those of the author(s) only and do not necessarily reflect those of the European Union or the European Research Executive Agency (REA). Neither the European Union nor the granting authority can be held responsible for them. The authors thank the financial support supplied by the grant PID2022-137061OB-C21 funded by MI-CIU/AEI/10.13039/501100011033 and by “ERDF/EU”. The authors also thank the funding supplied by the Consellería de Cultura, Educación, Formación Profesional e Universidades (accreditations ED431G 2023/01 and ED431C 2025/49) and the European Regional Development Fund, which acknowledges the CITIC, as a center accredited for excellence within the Galician University System and a member of the CIGUS Network, receives subsidies from the Department of Education, Science, Universities, and Vocational Training of the Xunta de Galicia. Additionally, it is co-financed by the EU through the FEDER Galicia 2021-27 operational program (Ref. ED431G 2023/01).

\bibliographystyle{elsarticle-num} 
\bibliography{biblio}

\end{document}